# Contextual Spelling Correction with Language Model for Low-resource Setting

Nishant Luitel[1], Nirajan Bekoju[2], Anand Kumar Sah[3] and Subarna Shakya[4]
*Dept. of Electronics and Computer Engineering,*
*Pulchowk Campus, Tribhuwan University,*
Lalitpur, Nepal
076bct041.nishant@pcampus.edu.np[1], 076bct039.nirajan@pcampus.edu.np[2], anand.sah@pcampus.edu.np[3], drss@ioe.edu.np[4]

*Abstract*—The task of Spell Correction(SC) in low-resource languages presents a significant challenge due to the availability of only a limited corpus of data and no annotated spelling correction datasets. To tackle these challenges a small-scale word-based transformer LM is trained to provide the SC model with contextual understanding. Further, the probabilistic error rules are extracted from the corpus in an unsupervised way to model the tendency of error happening(error model). Then the combination of LM and error model is used to develop the SC model through the well-known noisy channel framework. The effectiveness of this approach is demonstrated through experiments on the Nepali language where there is access to just an unprocessed corpus of textual data.

*Index Terms*—Noisy channel, Language model, Error model, Unsupervised

## I. Introduction

Accurate and reliable written communication is fundamental in various domains, from academia to professional settings, and its importance cannot be overstated. Spelling mistakes provide a serious problem since they can make communication and comprehension difficult. The task of spelling correction(SC), therefore, stands as an important component in natural language processing (NLP) systems.

Spelling errors generally manifest in two primary categories: real-word errors and non-word errors [1]. Real-word errors refer to instances where a misspelled word exists as a valid word in the language. These errors often result in the unintentional usage of a correctly spelled word that, unfortunately, does not convey the intended meaning in the given context. For example, 'हात धुनुहोस् र स्वास्थ जीवन जिउनुहोस् I' translates to wash your hands to stay healthy. However if the word 'हात' is changed to 'हार' which is within a single edit distance of the original word meaning 'hand', we get a real word that translates to 'a piece of jewelry which doesn't make sense contextually. Conversely, non-word errors involve misspellings that result in sequences of characters that do not constitute valid words in the language. A decent spelling correction model should be able to handle both types of errors.

In this paper, the statistical technique of noisy channels to perform spelling correction is focused. The term 'noisy channel' originates from information theory, where it refers to a communication channel that introduces random noise into transmitted signals. Instead of attempting to directly identify and correct errors in the observed text, this approach involves estimating the most likely intended message given the observed text and the probabilities of different error sources [2]. Key components of noisy channel spelling correction include language models, error models, and candidate generation.

As pointed out already, the correction of real word errors requires contextual understanding. Contextual understanding in a text can be captured using a Language model. Recent language models that are trained on a huge corpus of data are shown to have the ability to do tasks in a few-shot setting [3]. One could try using such LLMs to build context-sensitive spell checkers [4], [5]. However, training such huge LMs is expensive, and for low-resource languages, enough quality datasets are a bottleneck. A method of spelling correction that requires a minimal amount of resources is proposed and a decent small transformer-based LM on a corpus of only 1.2 GB data is trained. LMs provide probability that a given word occurs in the presence of its context. Hence, the list of candidate words can be ranked based on the contextual probability given by LMs.

Although language models provide good contextual understanding, error model is required that provides a higher weighting to candidate words that are likely to generate the error words. In other words, though some candidate words can be contextually appropriate they can still not be the intended word because they are unlikely to be typed as the observed noisy word. Hence, balanced weighting from both LMs and the error model is necessary. This is equivalent to performing Bayesian inference to determine the posterior distribution as described in section III.

A simple way to generate candidates for words includes searching a fixed vocabulary to find real words that are within a given edit distance. An edit distance of 2 is a decent choice. However, for smaller words, even with an edit distance of 1 can result in a large number of candidates which can be difficult to handle computationally. Hence, to reduce the number of candidates for every word, a simple method is developed as described in section III-A.

Due to the intrinsic difficulty of Nepali script as well as the influence of English medium communication, the quality of published Nepali text in the media has been decreasing. Hence, the Nepali language can benefit greatly from the use of automatic spelling correction systems. The main contributions of this paper are as follows:

1) An autoregressive language model based on word-based vocabulary with around 350,000 tokens is trained.
2) An error model is trained from a corpus of text in an unsupervised way and proposed a simple algorithm to add noise that resembles human-like errors. This can be used to create SC datasets.
3) Comparative study is performed based on character and word level accuracy using 2 error models and 5 variants of language models. Ablation study is performed to show the level of corrective behavior using only the error model. We have made our code open via GitHub[1].

## II. Related Works

Spelling correction (SC) systems in the Nepali language have predominantly been dictionary-based, relying on manual rules to generate comprehensive lists of words [6], [7]. These systems utilize the Hunspell framework. A different approach used includes using deep learning to perform end-to-end training [8]. However, other successful approaches such as the noisy channel model haven't been tried for Nepali which is explored in this paper.

SC generally involves the detection of error words, finding candidates for correction, ranking the candidates, and generating noisy datasets either for end-to-end training or evaluation. Detection of error is done mostly by recognizing out-of-vocabulary(OOV) words. However, recognizing only OOV words as errors don't lead to spell checking on real word errors [9]. Hence, to correct real word errors, every word is considered to be a likely noisy word and candidates are generated for each word. These candidates are generated either using Damerau-Levenshtein edit distance directly to error word [10] or by using edit distance on metaphones [9]. Authors in [11] use both DL edit distance of up to 2 and double metaphone to generate candidates of edit distance 3. Besides using edit distance, using language models to produce top probability candidates has been another popular choice [12], [13]. Nonetheless, this may not be an ideal choice if the language model itself has a noisy or misspelled vocabulary. Once the candidates are generated they have to be ranked. Noisy channel models rank the candidates based on full Bayesian inference i.e. combined weighting of the error model(likelihood) and language model(prior) [1], [2]. Error models that have been used include equal distribution over candidates [2], the inverse of edit-distance [9] and partition based approach developed in [14]. Similarly, language models used typically involve probabilistic n-grams [9], [15]. Though the revolution has been seen in language models trained using neural networks, direct probability estimates from such models haven't been used together with error models to calculate weightings for the candidates. We intend to do just that. One approach that also uses probability estimates from Neural LM is taken in [12], where BERT-based LM is used to determine the candidates for each word. They then use edit distance to find words with minimum edit distance among the candidates to perform the correction. The approach used here contrasts with this, as candidates are initially generated using edit distance and subsequently ranked using Neural LM and the channel model.

For low-resource languages, datasets for evaluating spelling correction models generally don't exist. Hence, a noisy dataset is generated using different methodologies. The simplest approach to generating errors is by introducing random edits (deletion, addition, substitution, and replacement) with some probability to characters but it doesn't generate realistic human-like errors [16]. Other approaches include replacing some of the words with their frequent misspellings [12], [17] and, finding and applying rules that generally occur in misspellings [16], [18]. Authors in [16] suggest translating confusion matrix in high resource language to low resource language via keyboard mapping, but it wouldn't be relevant for language like Nepali where misspellings occur due to intrinsic hardness in recognizing spellings rather than the action of mistyping. A large dataset of misspelled and corresponding correct sentences can also facilitate training end-to-end networks for SC systems [18], [19].

## III. Methodolody

### A. Candidate Generation

Considering all the candidates of observed words within a decent edit distance such as 2 can be quite expensive. To reduce the number of possible candidates, A simple heuristic is designed. Initially, candidates for words with a length of 3 or less is generated using an edit distance of 1 and for words with a length larger than 3 using an edit distance of 2 or less. If the candidate list has less than 5 words then we return the list. Otherwise, both the observed word and the words in the candidate list are transliterated to English and, then sort the pair in ascending order using edit distance calculated on the double metaphone as the key. Finally, only the candidates that have minimum edit distance with the observed word are returned. Damerau-Levenshtein edit distance is used in this approach.

What can be done with words only, can also be done for entire sentences. For this [1] suggests combining permutation of candidates of each word to form a complete list of candidate sentences. Similar to ranking words one can also rank entire candidate sentences. However, this is computationally more expensive than using words only. This is explained in further subsections.

### B. Training an Error Model

Training an error model amounts to learning $P(x|w)$ i.e. the probability that the observed noisy word 'x' is generated from the intended word 'w'. Authors in [14] approximate this likelihood by finding the partitions 'R' in observed words and partitions 'T' in candidate words that maximize it. Mathematically,

$$P(x|w) \approx \max_{R,T s.t. |R|=|T|} \prod_{i=1}^{|R|}[P(T_i|R_i)] \quad (1)$$

---

[1] https://github.com/NishantLuitel/Nepali-Spell-Checker

As seen in eq.1, observed words are partitioned into substrings and the substrings are aligned with the partitions of candidate words. To estimate, $P(T_i|R_i)$ simple counting-based probability can be calculated for all such substring alignments using triples of the correct word, misspelled word, and frequency extracted from the corpus in an unsupervised way. The creation of these triples, as demonstrated in [15], involves the extraction of terms with close edit distances. The underlying assumption is that correct words manifest approximately 10 times more frequently than their closely misspelled counterparts. In alignment with this methodology, the used approach involves the extraction of terms within an edit distance of 2, only considering those that exhibit a frequency fivefold greater than their misspelled counterparts within the analyzed corpus. We let the maximum length of each partition be a single character to keep things simple and, allow partitions to include empty string to allow deletion or addition.

Figure 1. is the heat map representing the edit probabilities of some devnagari characters(all not included due to space limitation) learned from training the error model. It captures a larger proportion of the probability for character substitutions that is likely for a human. For example 'ि' -> 'ी' and 'श' -> 'स' substitutions have redder color in the heatmap representing likely human errors. This heatmap of edit probabilities is used for generating errors. The one used for correction is shown in the appendix.

Additionally, to build an error model for candidate sentences, independence is assumed and hence naively multiply the error probability of individual words. For observed sentence '$X$' composed of words $x_1, ..., x_l$ with candidate correction sentence '$W$' composed of words $w_1, ..., w_l$, $P(X|W)$ is given by,

$$P(X|W) = \prod_{i=1}^{l}[P(x_i|w_i)] \quad (2)$$

*C. Training a Neural LM*

Recent State of the art LMs are trained either using bidirectional transformers [20], or with autoregressive transformers [21], [22]. Similar to probabilistic LMs, autoregressive neural LMs also provide the probability value of the next token given the context. However, neural models can more efficiently remember contexts than their probabilistic counterparts. In the scenario of contextual spelling correction, these models can provide us with a probability of occurring of each candidate word for the given context. Hence, the probability $P(w)$, of candidate word '$w$' to occur for contextual observed words $(x_{i-1}, x_{i-2}, ...)$ is given by,

$$P(w) = P(w|x_{i-1}, x_{i-2}, ...) \quad (3)$$

BERT-based LMs use future context alongside with past context. Hence, the probability $P(w)$ for such bidirectional transformers is given by,

$$P(w) = P(w|..., x_{i+1}, x_{i-1}, ...) \quad (4)$$

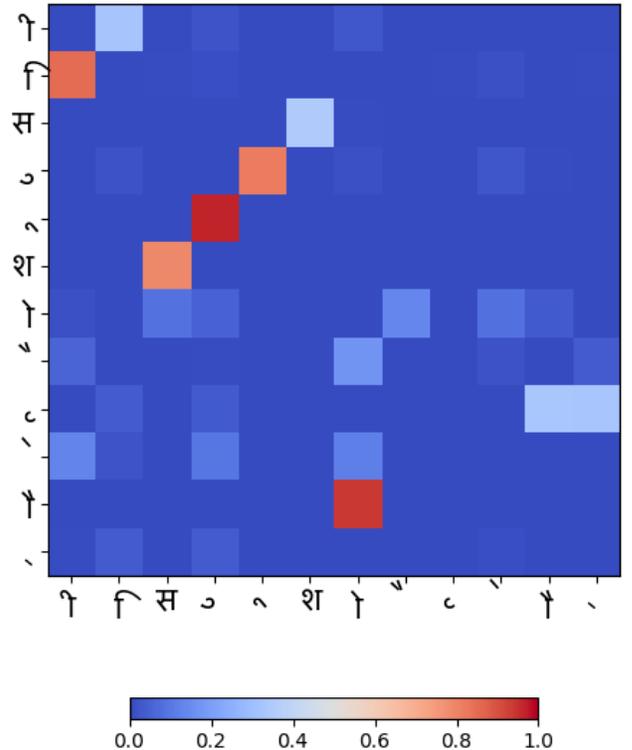

Figure 1: Heat map showing edit probability from the trained BM model for error generation

Similarly, to calculate the probability of candidate sentence '$W$' composed of individual words $(w_1, ..., w_l)$ to occur, we again assume independence and naively multiply the probability of occurrence of individual words. Hence,

$$P(W) = \prod_{i=1}^{l}[p(w_i)] \quad (5)$$

For this research, word vocabulary-based autoregressive neural transformer is trained using cross-entropy loss which allows to work directly with word-level tokenization. It is also compared with BERT-based language models that are openly available [23], [24].

*D. Combining Error Model and LM*

To rank the candidates, the posterior distribution '$P(w|x)$' is calculated by performing full Bayesian inference using Bayes theorem:

$$P(w|x) = \frac{P(x|w)P(w)}{P(x)} \quad (6)$$

Here, the likelihood term corresponds to the error model as defined before and prior as the language model accordingly. The goal is to find the word '$w$' that maximizes eq.6. Since $P(x)$ is constant for any '$w$', it is safe to ignore the denominator and find the most likely intended word as:

$$\hat{w} = \arg\max_{w} P(x|w)P(w) \quad (7)$$

Further, a hyperparameter '$\lambda$' is defined to provide a greater degree of freedom in eq. 7. The value of '$\lambda$' is estimated by inspecting its result in a few sentences. Hence the final equation becomes,

$$\hat{w} = \arg\max_{w} P(x|w)P(w)^{\lambda} \quad (8)$$

Accordingly, Bayesian inference is applied to rank the candidate sentences as well.

$$\hat{W} = \arg\max_{W} P(X|W)P(W)^{\lambda} \quad (9)$$

### E. Evaluation Dataset

Around 320 Nepali sentences is gathered and then used the error model trained in sec. III-B to generate errors in them. Algorithm 1. shows the pseudocode for this process. First, *None* token 'n' is added between each character in every word but only if both preceding and successive character isn't a character modifier. Character modefiers in Nepali includes characters 'ि','ी','ो','ौ','ु','े','ै','ा','ृ','ॄ','ॢ' and 'ा'. Then, for each character in a *None* added word, *with_probability* function is used to return either **True** or **False** with probability proportional to the frequency of that character in our error model as returned by the function *count_probab*. Finally, if the corresponding character has returned **True**, we look at that character's dictionary with probabilities to get replaced with other characters(returned by *dict_selector* function) and sample a character using that distribution.

---

**Algorithm 1:** Pseudocode for error generating function with python-like syntax

**Data:** tokens (set of words in a sentence)
(1) **for** $i \leftarrow 0$ **to** $len(tokens) - 1$ **do**
(2)    **for** $j, t$ **to** $enumerate(tokens[i])$ **do**
(3)       $temp \leftarrow temp + t$
(4)       **if** $j == len(tokens[i]) - 1$ **then**
(5)          **continue**;
(6)       **if** $t \notin NepaliCharacterModifiers$ **and** $tokens[i][j+1] \notin NepaliCharacterModifiers$ **then**
(7)          $temp \leftarrow temp + $ 'n'
(8)    $result \leftarrow [with\_probability(count\_probab[t])$ **if** $t \in dict\_selector$ **else False for** $t$ **in** $temp]$
(9)    $li \leftarrow [ch$ **if** $(b \neq True$ **or** $ch \notin dict\_selector)$ **else** $sample\_from\_probability(dict\_selector[ch])$ **for** $ch, b$ **in** $zip(temp, result)]$
(10)    $tokens[i] \leftarrow$ ''.join([ '' **if** $ch$ is None **or** $ch ==$ 'n' **else** $ch$ **for** $ch$ **in** $li]$)
(11) **return** $tokens$

---

To make a fair comparison of this method, two error models are trained on two different corpora. Error model trained from the dataset 'A large scale Nepali text Corpus' [25] was used for generating the evaluation dataset whereas, the model trained from 'Oscar' dataset [26] was used during actual evaluation.

## IV. Experimental Setup

We try out the combination of two error models and five language models to evaluate the generated dataset.

### A. Error Models

1) **Constant distributive(CD)**: In this error model, '$\alpha$' parameter is set which represents the probability that the observed word is correct i.e. $P(w|w)$ then, we uniformly distribute '$1 - \alpha$' over all the candidates of '$x$' except '$w$' [2]. Mathematically,
The absolute value of $x$ is defined

$$P(x|w) = \begin{cases} \alpha & \text{if } x = w \\ \frac{1-\alpha}{|C(x)|-1} & \text{if } x \in C(x) \\ 0 & \text{if } x \notin C(x) \end{cases}$$

We choose the value of $\alpha$ to be 0.65.

2) **Brill and Moore(BM)**: BM error model was trained as described in section III-B with two corpus of data. Training performed in the 'Oscar' dataset [26] was used for evaluation.

### B. Language Models

1) **KnLM-2**: Probabilistic bigram language model is trained using [26] dataset with Kneser-Ney smoothing which allows backing off to unigrams when bigram isn't available. Inference using n-gram language models are fast because they involve dictionary lookup. However, the memory requirement increases drastically with increasing 'n' so it is not efficient to capture longer contexts. We use candidate sentence ranking with this model.

2) **NepaliBert**: NepaliBert was trained by [23] in Masked Language modeling fashion using the BERT architecture described in [20]. Candidate sentence ranking is used with this model.

3) **deBerta**: deBerta is the Bert-based Nepali LM trained by [24] following the approach used in [27]. Basically, instead of using the sum of positional encoding with the embedding of tokens, deBerta model uses a separate embedding to positional encoding as well and uses the disentangled attention calculated from both relative position and content. Like NepaliBert, we again use candidate sentence ranking.

4) **Trans**: We have trained an autoregressive LM with 4 transformer encoders and the same number of decoders on the word-based vocabulary of around 350,000 using the Oscar corpus [26]. An embedding size of 300 was used with four attention heads in the transformer. The dropout value was 0.05. It was trained with cross-entropy loss. The trans model also uses candidate sentence correction.

5) **Trans-word**: This is the same language model as described in 4. However, it uses candidate word correction rather than candidate sentences.

## C. Evaluation Metrics

SC models are evaluated and compared based on the following metrics:

1) **Word Error Rate(WER)**: WER is the ratio of the total number of word-level errors to the total number of words in the reference text. A lower WER indicates better performance in terms of word-level accuracy.
2) **Word Accuracy**: Word accuracy is computed by dividing the number of accurately corrected words by the total number of error words present in the text. A higher accuracy indicates a better SC model.
3) **Character Accuracy**: To calculate character accuracy, alignment is created between each word in the generated dataset and the corresponding ground truth word. Then the ratio taken between all the corrected characters to the total error character represents character accuracy.

## V. Result

Tables I, II, and III show WER, word accuracy, and character accuracy respectively for two error models and 5 variants of language models used as described in section IV.

WER calculation is found to be 0.25 when no correction was applied. When this score is compared after correction with different models, it is observed that the Trans version decreases the WER to lowest among other models. Bert-based models introduce more error after correction suggesting overcorrection. WER score is lower for BM models compared to CD models except for NepaliBert. The correct label dataset is itself noisy, hence even the accurate corrections for some cases have increased the WER. The only way to get rid of this noise is to manualy correct the misspellings(not performed here) in the original dataset considered as the correct label dataset.

In terms of word accuracy, Trans with BM is best performing with an accuracy of about 69.1%. The performance of probabilistic LM(Knlm-2) and Trans-word are comparable. Again the Bert-based models are heavily outperformed by other models. CD models tend to be better than BM models in terms of character accuracy. We note that character accuracy is higher than word accuracy for corresponding models. This means that there are some error words where few but not all of the characters are corrected.

Although Bert-based language models don't match the performance of other models, it doesn't mean that they are less capable as a language model. It probably means that our method of correction doesn't work well with BERT-like models.

Table I: Word error rate(WER) after correction.

|    | Knlm-2 | deBerta | NepaliBert | Trans | Trans-word |
|----|--------|---------|------------|-------|------------|
| CD | 0.201  | 0.372   | 0.321      | 0.145 | 0.168      |
| BM | 0.187  | 0.299   | 0.248      | 0.119 | 0.166      |

Table II: Word accuracy

|    | Knlm-2 | deBerta | NepaliBert | Trans | Trans-word |
|----|--------|---------|------------|-------|------------|
| CD | 0.601  | 0.403   | 0.384      | 0.665 | 0.599      |
| BM | 0.647  | 0.439   | 0.422      | 0.691 | 0.647      |

Table III: Character accuracy

|    | Knlm-2 | deBerta | NepaliBert | Trans | Trans-word |
|----|--------|---------|------------|-------|------------|
| CD | 0.671  | 0.535   | 0.492      | 0.744 | 0.694      |
| BM | 0.683  | 0.495   | 0.443      | 0.690 | 0.695      |

## VI. Ablation study

Ablation study is performed by removing the language model entirely from the probability calculations. The objective was to observe and compare whether the error model by itself has some corrective abilities. From the Bayesian perspective, the prior is constant which amounts to performing a maximum likelihood estimate with possible candidates. Hence, the candidate word, that is most likely to result in the given error ignoring how likely it is to occur within the context, is chosen. However, the error model will always return a larger probability score for actually observed words even if they are not intended so we have chosen to use a correctly spelled vocabulary list of around 125000 words and to provide a larger negative score if the observed word isn't in the vocabulary. This way error model compares the words that are in the vocab to find a maximum likelihood candidate. The vocabulary list was formed by extracting Nepali words from the Nepali dictionary using OCR.

Table IV: Word Candidate based(Ablation study)

|    | WER   | Word acc. | Char acc. |
|----|-------|-----------|-----------|
| CD | 0.286 | 0.395     | 0.532     |
| BM | 0.186 | 0.457     | 0.485     |

Table V: Sentence Candidate based(Ablation study)

|    | WER   | Word acc. | Char acc. |
|----|-------|-----------|-----------|
| CD | 0.286 | 0.395     | 0.532     |
| BM | 0.188 | 0.371     | 0.358     |

The result shown in table IV is the performance of the SC model where word-based correction is used. It can be observed that the WER has dropped from 0.25 to 0.186 for the BM model. However, for the CD model, the WER has increased to 0.286 showing an overcorrecting nature. Hence, it can be concluded from this analysis that the BM model has a better correcting ability than the CD model. However, the character accuracy for CD is higher than that of BM. Similarly, the result shown in table V uses sentence-based correction. This was analyzed only for 200 sentences. Again, it can be observed that the BM model better captures the errors than CD in terms of word accuracy and WER but fails to outperform on character accuracy.

## VII. Scope and Limitations

SC systems find many applications both in assisting humans to write correctly as well as within the context of Machine Learning. SC can also be employed to preprocess the input text to tackle adversarial attacks with misspellings [17], [28].

It can be used to create higher-quality NLP datasets. Moreover, it can also be applied to correct misspelled tokens by post-processing outputs from applications such as OCR and ASR [29].

One limitation of our method is that it only handles the cases where input tokens are the same number as the correct label tokens. Hence, two tokens 'म लाई' is not corrected to the correct token 'मलाई'. Additionally, an error model have been employed that learns human-like errors from the corpus, which is not suitable for usage in situations where the error-generating distribution is much different, such as correcting OCR predictions. Similar can be said about the post-correction of ASR system outputs. Nonetheless, it is hypothesized that the method proposed might exhibit some corrective behavior even in those situations due to reasonable candidate creation.

## VIII. Future Work

Future direction that can be taken is to train the model end to end using seq2seq architectures in subword or word level using error generation process to create the larger datasets. In the context of the noisy channel method, this is equivalent to learning both the error model and prior jointly with the same network. Additionally, introducing errors by splitting a correct word or joining two consecutive separate words can be used to handle the cases of different length input and correct label tokens.

Further, enhancement can be done to the model by introducing NER to correctly recognize and ignore correcting name entities in the sentence.

## IX. Conclusion

Although neural architectures have been shown to have great generalizing ability, they require large amounts of data. For low-resource languages such as Nepali, one has to rely on unsupervised methods to tackle problems such as SC. In this paper, it is shown with just a single noisy corpus of text, one can train both an error model and, an LM using SOTA techniques to create a decent SC system using a noisy channel approach. Furthermore, this approach can be adapted to any language.

## Acknowledgment

We are extremely thankful to all the researchers for making open the language models [23], [24] and the datasets [25], [26] used in this paper.

# Appendix

## A. Error Model Edit Probability

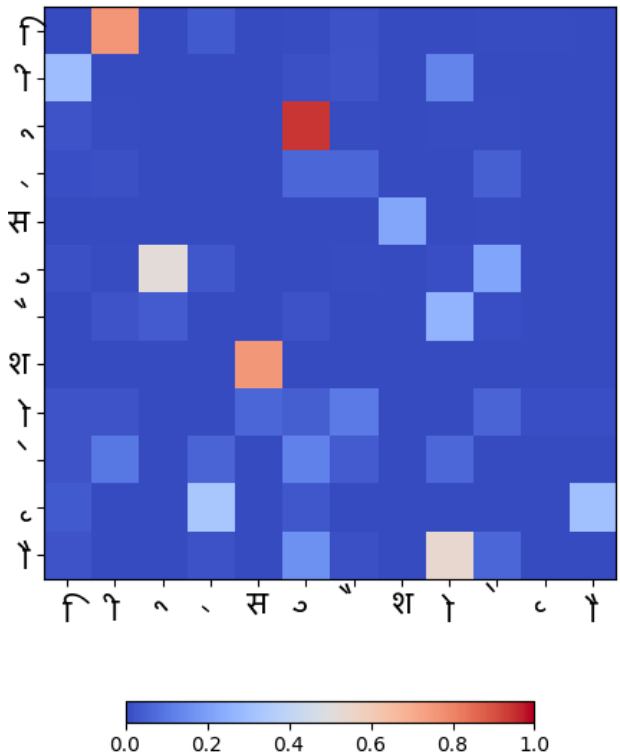

Figure 2: Heat map showing edit probability from the trained BM model used for correction

The heatmap in Figure 2 shows edit probabilities of some devanagari tokens(all not shown due to space limitation) trained from the Oscar data corpus. This model is used for performing correction.

## B. Training Data Analysis

Two different corpora have been used to train two separate BM models. Hence, comparative analysis of these two corpus using histograms was performed as shown in Figure 3-5.

In Figure 3, for better visualization of the distribution plot, for the IEEE dataset, the distribution plot is shown for only those paragraphs that have no. of words less than 80, and for the Oscar dataset, the number of words is less than 150.

Table VI: words per paragraph summary

| Dataset | 1st Quartile | 2nd Quartile | 3rd Quartile | Mean | Maximum |
|---|---|---|---|---|---|
| IEEE | 2 | 19 | 47 | 34.18 | 11501 |
| Oscar | 31 | 59 | 210 | 232.78 | 17904 |

In Figure 4, for better visualization of the distribution plot, for the IEEE dataset, the distribution plot is shown for only those paragraphs with no. of sentences less than 20, and for the Oscar dataset, the number of sentences less than 50.

Table VII: sentence per paragraph summary

| Dataset | 1st Quartile | 2nd Quartile | 3rd Quartile | Mean | Maximum |
|---|---|---|---|---|---|
| IEEE | 1 | 2 | 4 | 3.0796 | 606 |
| Oscar | 2 | 4 | 13 | 14.38 | 1060 |

From the descriptive statistic shown in table VI, we can see that the Oscar dataset has a large no. of words per paragraph compared to the IEEE dataset. Finally, for different cut-off frequencies, the no. of vocab obtained is shown for the Oscar and IEEE datasets in Figure 5.

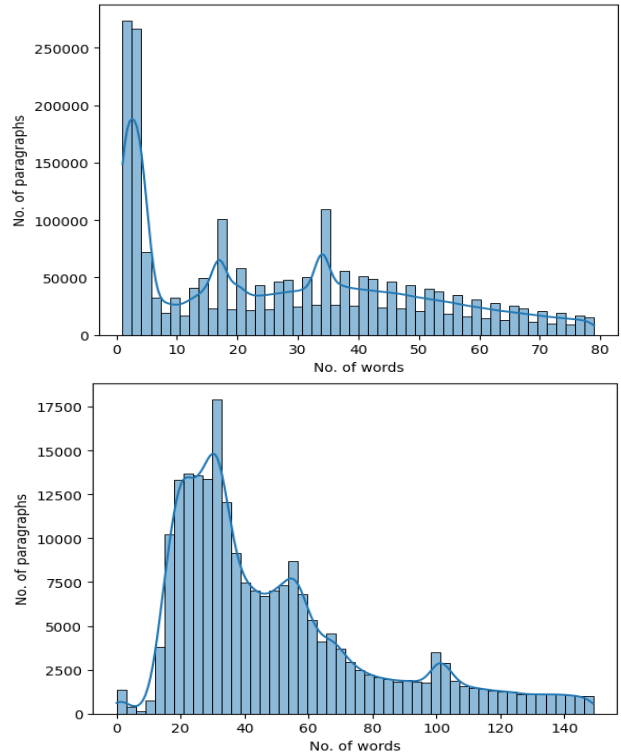

Figure 3: IEEE(top) vs Oscar(bottom) corpus words per paragraph analysis

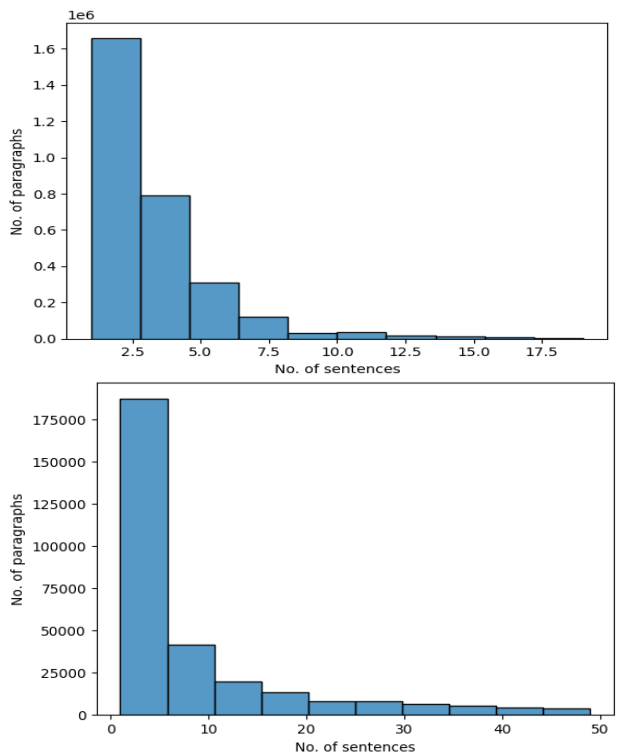

Figure 4: IEEE(top) vs Oscar(bottom) corpus sentence per paragraph analysis

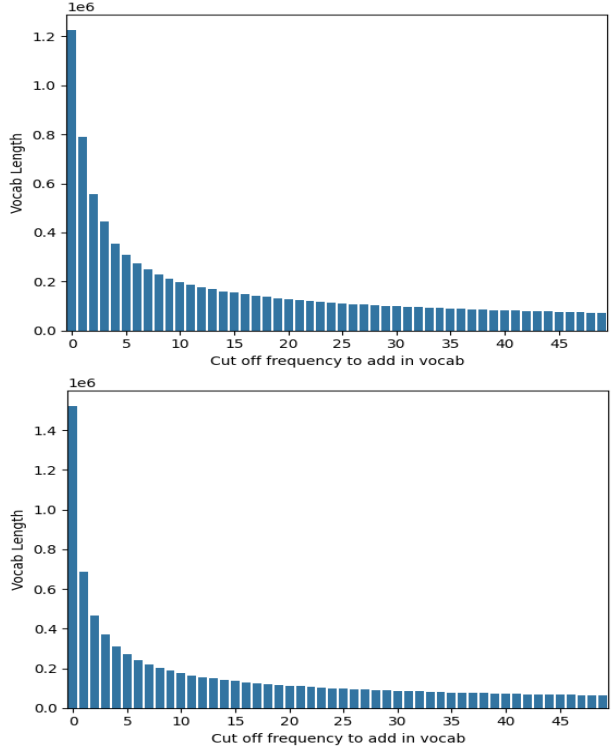

Figure 5: IEEE(top) vs Oscar(bottom) corpus vocab length for given cut-off frequency analysis